\documentclass[9pt,conference]{IEEEtran}
\usepackage{ifpdf}

\usepackage{cite}

\ifCLASSINFOpdf
   \usepackage[pdftex]{graphicx}
   \graphicspath{{images/}}
   \DeclareGraphicsExtensions{.pdf,.jpeg,.png}
\else
   \usepackage[dvips]{graphicx}
   \graphicspath{{images/}}
   \DeclareGraphicsExtensions{.eps}
\fi
\usepackage{amsmath}
\ifCLASSOPTIONcompsoc
  \usepackage[caption=false,font=normalsize,labelfont=sf,textfont=sf]{subfig}
\else
  \usepackage[caption=false,font=footnotesize]{subfig}
\fi

\usepackage{stfloats}
\usepackage{url}

\usepackage{multicol}
\usepackage{multirow}
\usepackage{amssymb}
\usepackage{colortbl}
\usepackage{tabularx}
\usepackage{lscape}
\usepackage{pdflscape}
\usepackage{booktabs}
\usepackage{lipsum}
\usepackage{gensymb}
\usepackage[acronym,nomain,nonumberlist]{glossaries}
\usepackage{subfig}
\definecolor{row1}{rgb}{0.95,0.95,0.95}
\definecolor{row2}{rgb}{0.85,0.85,0.85}
\newacronym{5g}{$5G$}{$5^{th}$ Generation Mobile Network} 
\newacronym{cacc}{CACC}{Cooperative Adaptive Cruise Control}
\newacronym{etsi}{ETSI}{European Telecommunications Standards Institute}
\newacronym{it}{IT}{Information Technologies}
\newacronym{its}{ITS}{Intelligent Transport System} 
\newacronym{iot}{IoT}{Internet of Things}
\newacronym{mec}{MEC}{Multi-Access Edge Computing}
\newacronym{ran}{RAN}{Radio Access Network}
\newacronym{sdn}{SDN}{Software Defined Networking}
\newacronym{v2x}{V2X}{Vehicle-to-Everything}
\newacronym{5gaa}{$5$GAA}{$5G$ Automotive Association}
\newacronym{json}{JSON}{JavaScript Object Notation}
\newacronym{tcp}{TCP}{Transmission Control Protocol}
\newacronym{rud}{RUD}{Road User Description}
\newacronym{id}{ID}{Identification}
\newacronym{rmse}{RMSE}{Root-Mean-Square Error}
\newacronym{gui}{GUI}{Graphical User Interface}
\newacronym{gdm}{GDM}{Global Dynamic Map}
\newacronym{uuid}{UUID}{Universally Unique Identifier}

\begin{document}
%
\title{A Lane Merge Coordination Model for a V2X Scenario}
%
%
%

\author{\IEEEauthorblockN{Luis Sequeira, Adam Szefer, Jamie Slome and Toktam Mahmoodi~\IEEEmembership{Senior~Member,~IEEE}
\thanks{L. Sequeira, A. Szefer, J. Slome and T. Mahmoodi are with the Department of Informatics at King's College London. Email: \{luis.sequeira, adam.szefer, jamie.slome, toktam.mahmoodi\}kcl.ac.uk.}
}

\IEEEauthorblockA{Centre for Telecommunications Research, Department of Informatics,  King's College London, London, UK\\
Email: \{luis.sequeira, adam.szefer, jamie.slome, toktam.mahmoodi\}@kcl.ac.uk}

}

%
%

\markboth{Journal of \LaTeX\ Class Files,~Vol.~14, No.~8, August~2015}%
{Shell \MakeLowercase{\textit{et al.}}: Bare Demo of IEEEtran.cls for IEEE Journals}
%



\maketitle

\begin{abstract}

Cooperative driving using connectivity services has been a promising avenue for autonomous vehicles, with the low latency and further reliability support provided by \gls*{5g}. In this paper, we present an application for lane merge co-ordination based on a centralised system, for connected cars. This application delivers trajectory recommendations to the connected vehicles on the road. The application comprises of a \textit{Traffic Orchestrator} as the main component. We apply machine learning and data analysis to predict whether a connected vehicle can successfully complete the cooperative manoeuvre of a lane merge. Furthermore, the acceleration and heading parameters that are necessary for the completion of a safe merge are elaborated. The results demonstrate the performance of several existing algorithms and how their main parameters were selected to avoid over-fitting.

\end{abstract}


%
\IEEEpeerreviewmaketitle

\section{Introduction} 
\label{Introduction}
Connected vehicles play a crucial role in the \gls{its}. \gls{its} is a platform capable of generating rich data relating to vehicles' functioning and their environment. In this domain, associations such as the \gls{etsi} and \gls{5gaa} have promoted the use of cellular \gls{v2x} communications.  A \gls{v2x} approach seeks to provide real-time, and highly reliable information to researchers and stakeholders. This has resulted in enhancements to road safety, traffic efficiency, environmental issues and energy costs \cite{its4}. 

Due to the growing scope of \gls{v2x}, a plethora of use-cases and applications are under research and development \cite{its2}: automated overtake, co-operative collision avoidance, high density platooning and lane merging. A vehicle capable of transmitting and receiving data from a network is likely to increase the awareness of a driving agent. On-board information can be transmitted across a variety of channels. This includes vehicle-to-vehicle and vehicle-to-network. These methods deliver the necessary functions to perform manoeuvres in a multitude of traffic situations.

In this paper, we focus on a lane merge scenario. This involves a vehicle merging onto a carriageway. We present a coordination model that uses a centralised system. This system delivers trajectory recommendations to connected vehicles. These recommendations account for all surrounding vehicles - connected or unconnected. For the calculations, time-critical variables include location, speed and acceleration. We evaluate various machine learning algorithms to predict whether a merging vehicle can execute the manoeuver safely. The contributions presented in this paper include:

\begin{itemize}
    \item A \textit{Traffic Orchestrator} model based on a centralised system, that delivers trajectory recommendations to connected vehicles.
    \item A performance evaluation of different machine learning algorithms to appropriately select one for the model.
    \item A study on the effect of maximum depth and the number of estimators for each algorithm.
\end{itemize}

The remainder of this paper is organised in the following way. Section \ref{State of the art} provides a state of the art of different approaches for lane merge algorithms. Section \ref{Architecture and System Model} presents the general system model. Section \ref{Traffic Orchestrator} demonstrates the purpose of the \textit{Traffic Orchestrator} model. Furthermore, the estimations of different machine learning algorithms are presented and analysed in Section \ref{Results}. Finally, a conclusion and future works are presented in Section \ref{Conclusion}.

\section{State of the art}
\label{State of the art}
Vehicles have gained more capabilities including cruise control, lane following and assisted large-lateral control manoeuvres. Nowadays, research is progressing towards more relevant topics like \gls*{cacc}, for lane changing and merging \cite{lm1}. To perform a safe merge, a safety distance is required. This distance is between the merging vehicle and other vehicles. If a merging vehicle struggles to fit into the existing gap between vehicles, there are two options.  The first option is to slow down or halt. By halting, the vehicle will avoid a collision. Alternatively, vehicles in the main lane could slow down, speed up or merge onto a third lane. Therefore, a lane merging algorithm is required. This will perform actions on a merging vehicle - executing successful and safe lane merges \cite{lm8}.

In \cite{lm2}, the authors simulated a lane merge scenario and applied pattern recognition for decision-making. The pattern model consists of a nine grid cell, in which each cell is marked as blocked or unblocked according to the information of surrounding vehicles. The longitudinal and lateral trajectory of the merging vehicle was fitted with a $ 5 \degree $ polynomial function. The simulation provides different trajectory models using active (accelerate/decelerate) and passive (wait) information.

There are function models to evaluate the decisions in a lane merging scenario. In \cite{lm3}, a low-complexity lane merging algorithm is presented. It determines whether a lane manoeuvre is desirable. If so, a suitable gap is selected and the time to perform the manoeuvre is given. The mathematical model calculates longitudinal and lateral control trajectories where several weighting parameters define the optimal model behaviour. However, the author has highlighted the need for a dynamic prediction model and the generation of backup trajectories. 

A similar approach was used in \cite{lm10}, where the idea was to reduce the additional road space before the lane change occurs. With this view, a two-lane road was divided into cells, that can be empty or contain a vehicle. Four different actions were proposed to manage vehicles (i.e, acceleration, slowing down, randomization, and vehicle motion). Based on this, three types of lane change were investigated: \textit{tail to head}, \textit{head to tail} and \textit{random}. The \textit{tail to head} approach showed better performance compared to a random lane merge. The algorithm assumes that the time and space for the vehicle to change lane is sufficient and all vehicles on the road are connected. In a more realistic scenario, time and distance between adjacent vehicles are essential in deciding if a lane change is possible.

Other strategies use on-board sensors. A representation of the road environment, using a Dynamic Probabilistic Drivability Map, is presented in \cite{lm9}, to provide adjacent lane merging instructions. The automotive test bed includes cameras, radars and lidar sensors. This delivers driving assistance based on a cost-sensitive analysis of the road environment, making use of dynamic programming. The theoretical formulation of this work was tested using data from $40$ real world merges.

Real world information has been used, but some of these works are at a primal stage. In \cite{lm7}, a work-in-progress for an on-ramp merge driving policy was presented. The scenario considers an on-ramp merge involving three vehicles - the merging vehicle and two vehicles on the main lane. A total of $9$ variables are used, where 5 variables describe the merging vehicle's driving state. These include speed, position, heading angle, distance to the right lane and distance to the left lane. For the other two vehicles, the speed and position are known. The algorithm has not been verified or validated.

In \cite{1m12}, the authors consider a transition stage in the path to fully autonomous transport, with mixed-autonomy driving. Their approach is based on a set of selfish factors. In the study, they consider mixed-autonomy driving as a collaboration, to ensure that the collective reward for lane merging is optimal. A thought-provoking part of the project introduces a user study. In this study participants simulated the role of a driver via a keyboard.

\begin{figure*}[!t]
    \centering
    \vspace{0.02in}
    \includegraphics[width=0.95\textwidth,height=6cm]{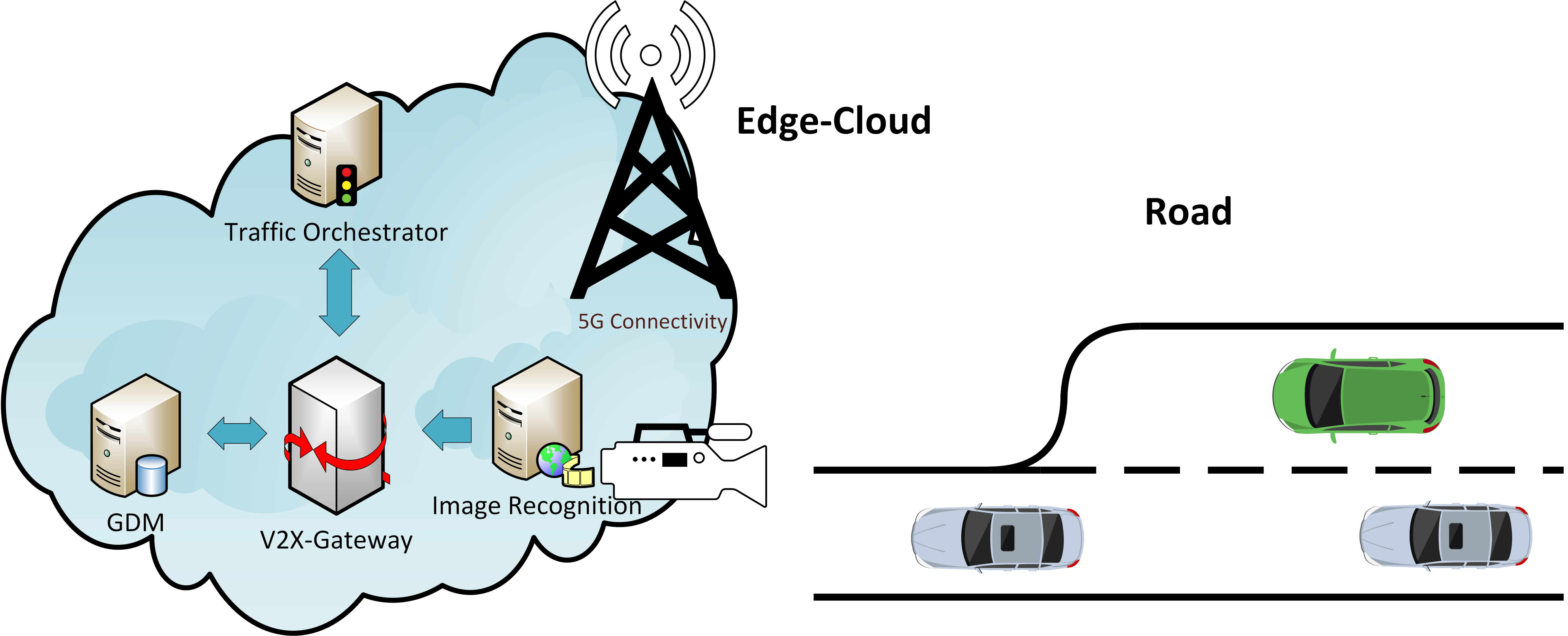}
    \caption{Lane merge coordination scenario.}
    \label{fig:LaneMergeCoordination}
\end{figure*}

\section{Architecture and System Model} 
\label{Architecture and System Model}
This section provides the architecture and model for a centralised coordination system. This system plans the trajectories of connected vehicles on the road to ensure there is sufficient space for a merging vehicle. The lane merge scenario examined in this work is depicted in Fig. \ref{fig:LaneMergeCoordination}. A connected vehicle will attempt to merge onto a single or multi-lane carriageway in which connected and unconnected vehicles are present. Through an edge-cloud approach, tailored trajectory recommendations are determined and sent by the merge coordination algorithm to connected vehicles. Four distinct components facilitate the lane merge coordination: a \textit{V2X-Gateway}, an \textit{Image recognition} system, a \textit{\gls{gdm}} and a \textit{Traffic Orchestrator}.

The \textit{V2X-Gateway} is responsible for forwarding messages to the appropriate applications and interfaces in the infrastructure. The \textit{V2X-Gateway} acts as a communication medium that connects the interfaces and applications to connected vehicles based on a message exchanging approach. This method of communication occurs across \textit{\gls{5g} connectivity}. To use the message forwarding functionality, applications must subscribe for access to trajectory and location information. The \textit{\gls{5g} connectivity} seeks to maintain a set of baseline requirements. The uplink per vehicle should be, at least, $320kbps$. Furthermore, the downlink per vehicle\footnote{\url{https://5gcar.eu/}} should be, at least, $4.7Mbps$. The end-to-end latency requirement should not exceed $30ms$.

An \textit{Image recognition} system \cite{kai} collects information about all the vehicles on the single or multi-lane carriageway. This information includes the localisation and trajectory-based parameters attributed to a specific road user (named, \gls{rud}). Regardless of whether a road user is connected or unconnected, information about that vehicle will be collected and processed. With trajectory-based information about all visible vehicles, the \textit{Image recognition} system communicates with a \textit{V2X-Gateway}. This will forward every message to the \textit{\gls{gdm}}.

The \textit{\gls{gdm}} stores environmental information about connected and unconnected vehicles in a database. It interprets messages containing vehicle information (\gls{rud}). This information is delivered by the \textit{Image recognition} system and connected vehicles. The \textit{\gls{gdm}} ensures that stored \glspl{rud} are updated. It also provides a synchronisation mechanism for descriptions originating from different sources (e.g., two \glspl{rud} from the \textit{Image recognition} system and a connected vehicle in a closely localised time frame, respectively).  To access information stored in the \textit{\gls{gdm}}, applications must subscribe to a specific location boundary.

The \textit{Traffic Orchestrator} will process environmental data about connected and unconnected vehicles to generate trajectories for connected vehicles. The \textit{Traffic Orchestrator} needs to consider time-critical variables such as the timestamp of the vehicle location, the speed of the vehicle and the vehicle-specific dimensions. Once the \textit{Traffic Orchestrator} provides a trajectory recommendation for a single or set of road users, these trajectories will be sent to the \textit{V2X-Gateway}. The \textit{V2X-Gateway} will feed these recommendations to the connected vehicles. The connected vehicles have the choice to either accept, reject or abort the recommendation. This feedback information is supplied by the connected vehicles to the \textit{Traffic Orchestrator}. Then, the feedback can be used to recalculate trajectory recommendations. The \textit{Traffic Orchestrator} will use data supplied to the \textit{\gls{gdm}} to accommodate a merge for an ``approaching vehicle". This results in a vehicle moving onto the carriage-way lane from an ``on-ramp" lane. In this paper, we will focus on the design of the \textit{Traffic Orchestrator}.

\section{Traffic Orchestrator}
\label{Traffic Orchestrator}
\subsection{Traffic Orchestrator Model}

\begin{figure}[!t]
	\centering
	\vspace{0.02in}
	\includegraphics[width=0.9\columnwidth,height=8.5cm]{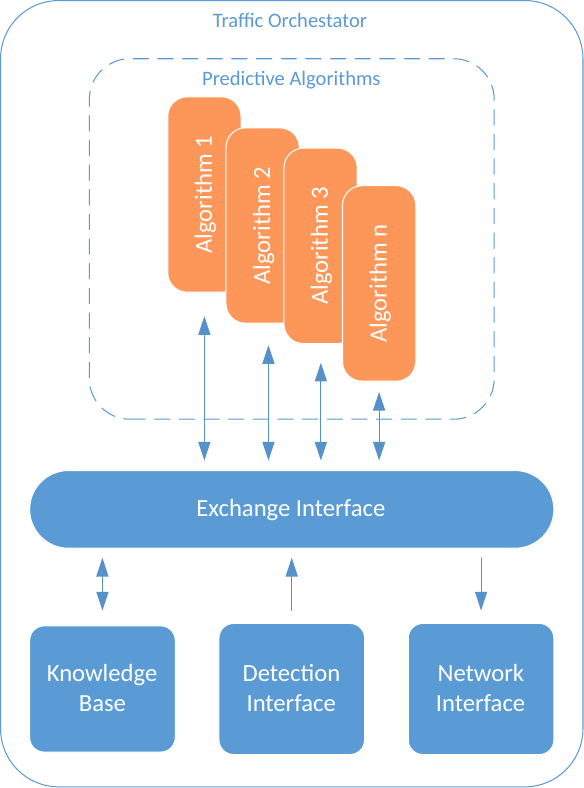}
	\caption{Proposed \textit{Traffic Orchestrator} architecture.}
	\label{proposed_architecture}
\end{figure}

The proposed architecture for the \textit{Traffic Orchestrator} system is presented in Fig. \ref{proposed_architecture}. The main purpose of the \textit{Detection Interface} is to listen to, wait for and receive any data being sent, over a \gls{tcp} connection, by the \textit{V2X-Gateway}. The \textit{Detection Interface} also acts as an intermediate filter that will read \gls{json} strings and commence the process of converting the \gls{json} readable messages into more compact and computer efficient entities. Once a message has been successfully parsed by the \textit{Detection Interface}, it can begin assigning the values found within the string to structured types. These structured types are then taken through further processing and data handling. Similarly, the \textit{Network Interface} will act as an intermediate filter that will convert and translate information within the \textit{Traffic Orchestrator}, to information that is readable and accepted by the \textit{V2X-Gateway}. 

A \textit{Knowledge Base} has been designed to store the information sent to the \textit{Traffic Orchestrator}. A knowledge base, of up-to-date \glspl{rud}, is maintained to guarantee that a manoeuvre recommendation is calculated based on all current road-environment knowledge. The \textit{Knowledge Base} will contain only the \glspl{rud} that the \textit{\gls{gdm}} has most recently transmitted. This will prevent maintaining information within the \textit{Traffic Orchestrator} that is out-of-date or no longer relevant. The \textit{Knowledge Base}, although simpler than a database, will have to mimic the access and modification functions of a typical database. 

The \textit{Exchange Interface} has been designed to have two responsibilities: execute the \textit{Traffic Orchestrator} application and mediate the flow of information across all interfaces in the \textit{Traffic Orchestrator}. The \textit{Exchange Interface} takes structured data from the \textit{Detection Interface} and appropriately forms the data into entities. These entities can then be reused throughout the rest of the system in a consistent manner. This component directly interfaces with the \textit{Knowledge Base} and will collect structured \glspl{rud}. Then, it will represent this data as a \textit{Traffic Orchestrator} entity. This entity will then be inserted into the \textit{Knowledge Base}. Another functionality of the \textit{Exchange Interface} is to provide access for consistent methods to a set of \textit{Traffic Orchestrator} functionalities, allowing different algorithms to run on top of it.

\subsection{Design premises for lane merge coordination} 

\subsubsection{Dataset}

Two distinct datasets collected by Federal Highway Administration Research and Technology\footnote{\url{https://www.fhwa.dot.gov}} are adopted in this work. The datasets represent the data collected from two American motorways: $I-80$ and $US-101$. In this paper, a scenario with $3$ cars is considered. For each lane change scenario, there is $1$ lane changing vehicle (denoted \textit{M}) and $2$ cars on an existing lane (denoted preceding - \textit{P} and following - \textit{F}).

\begin{figure*}[!t]
    \centering
    \vspace{0.02in}
    \begin{tabular}{c}
    
    \subfloat[\textit{true} recommendation.]{
        \includegraphics[width=0.3\textwidth]{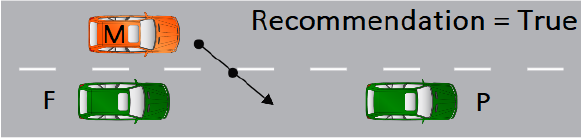}
        \label{fig:positiveRecommendation}
	}    
    \subfloat[\textit{false} recommendation.]{
        \includegraphics[width=0.3\textwidth]{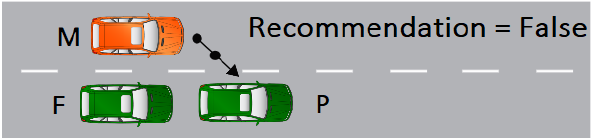}
        \label{fig:negativeRecommendation}
	} 
    \subfloat[Merging car is ``behind" following car.]{
        \includegraphics[width=0.3\textwidth]{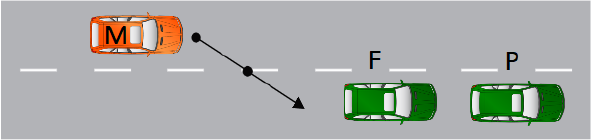}
        \label{fig:behindFollowing}
	}

    \end{tabular}
    \caption{Design premises for different road situations.}
    \label{fig:exp1-latency-vs-location}
\end{figure*}

\subsubsection{Detecting lane change}

Vehicle's measurements are sorted in ascending order by a timestamp. To detect a lane change, we need to compare two consecutive measurements. If the \gls*{id} value for a lane, associated with a vehicle, has changed, then a lane change has occurred. However, it is unknown whether the lane change is successful. To acquire a vast amount of data about a potential lane change, data from $4$ seconds before and $3$ seconds after the lane change is detected and subsequently stored ($70$ values per lane change). This provides information about locations on the road that should be deemed safe, or otherwise unsafe, for a lane change.

\subsubsection{Labelling the extracted data}
        
If a lane change is possible, the recommendation outputs \textit{true} (Fig. \ref{fig:positiveRecommendation}), otherwise \textit{false} (Fig. \ref{fig:negativeRecommendation}). This identifies which lane change situations are recommended and which are not. A lane merge is not relevant if the merging vehicle (M) is behind the following car (F) on the new lane. For this reason, all such cases are labeled with a recommendation equal to \textit{false} (Fig. \ref{fig:behindFollowing}). To change lane safely, a safe distance between the front of the merging vehicle and the back of the preceding vehicle (P) must be maintained between the vehicles. Furthermore, a safe distance between the back of the merging vehicle and the front of the following vehicle must be maintained between the vehicles. For both cases, a safe distance is considered to be $0.1$ times the speed of the merging vehicle.

\subsubsection{Desired position}

The most desirable position for the merging car is the location that fits the safety requirements for the merge. The preceding and following vehicles will determine the most suitable position on the merging lane.

\subsubsection{Recommended heading}

The heading for \textit{true} recommendations should ``lead" the vehicle to the position at which the car is most desired. For \textit{false} recommendations, the vehicle should be ``led" to the first location at which recommendations becomes \textit{true} (if possible). 

\subsubsection{Recommended acceleration}

The recommended acceleration is calculated with three rules in mind. Firstly, \textit{false} lane change recommendations use the average speed to the first point at which a recommendation is \textit{true}. Secondly, any position before the most desired position uses an average of the accelerations from the considered position, to the ``most desired position". Thirdly, for any position after the most desired location, an average of the accelerations from the ``most desired position" to the considered location, is calculated.

\subsection{Lane merge detection}

To detect a lane change, a script utilises previously prepared files with \gls*{json} objects - these objects contain a list of measurements for a specific car. The program stores the information about the initial lane number and compares that value with the current lane \gls*{id}. 

A recommendation checker was implemented to ensure that recommendations are accurate. For the checker to determine whether a recommendation could be \textit{true}, it checks if the gap is wide enough to accommodate a merging car. The minimal recommended gap value is dependable on the speed of the vehicle. 

The safety conditions of a lane merge are based on the intersection of two pairs of cycles. Firstly, one cycle has the central $x$ and $y$ coordinates of the merging vehicle and the radius of $10\%$ of its speed in $\frac{km}{h}$. This is compared with a second cycle which has the central $x$ and $y$ coordinates of the preceding vehicle and the radius of its length. If the cycles share common coordinates, or a cycle intersects the other, the recommendation is set to \textit{false}, resulting in a final recommendation for that case. Furthermore, the central $x$ and $y$ coordinates of the merging vehicle will be compared to a cycle with the central $x$ and $y$ coordinates of the following vehicle, with its radius equal to the length of the vehicle. Similarly, the same conditions are checked.

If the merging vehicle cycle does not intersect the cycles represented by the preceding or following vehicle, the recommendation is then set to \textit{true}, otherwise, it is set to \textit{false}.

\subsection{Acceleration and heading calculation}

The calculation for acceleration was divided into two segments - before arriving at the ``most desired position" (MSP) and after. Points before the MSP are calculated in reverse order (from the MSP to the first sample in the considered group). The accelerations for this measurement were calculated as a sum of the accelerations from the considered point to the MSP, divided by the number of samples between these two (constant time for each sample, if prefix sums were used). For the points after the MSP, the acceleration was calculated by dividing the prefix sum from the MSP to the considered point, divided by the number of samples between these two. 

Information about the MSP is used to calculate heading, but for the samples with \textit{false} recommendations, the MSP for the heading is changed to the first position at which the general recommendation is \textit{true}. 

\section{Results}
\label{Results}
\begin{figure*}[!t]
    \centering
    \begin{tabular}{c}
    
    \subfloat[Random Forest.]{
        \includegraphics[width=0.3\textwidth,height=4.5cm]{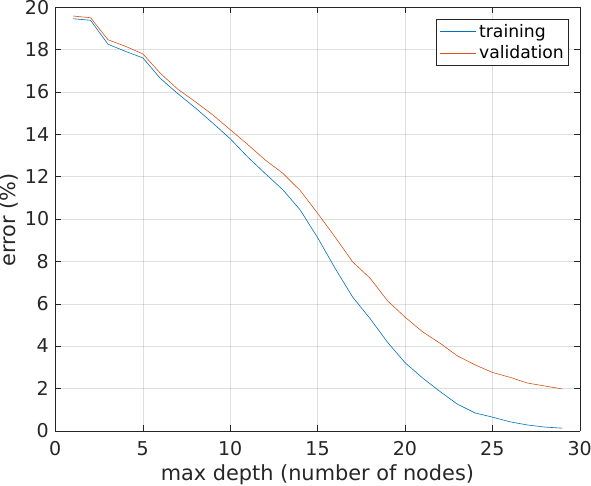}
        \label{fig:randomForestRecFeatures}
	}
        \subfloat[K-Nearest Neighbours.]{
        \includegraphics[width=0.3\textwidth,height=4.5cm]{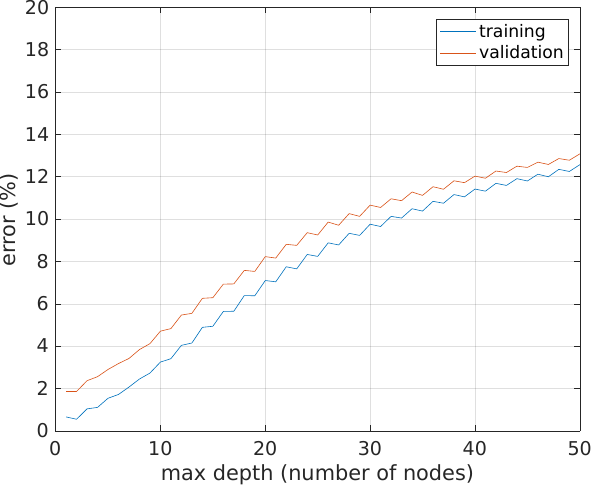}
        \label{fig:knnClass}
	}    
    \subfloat[Decision Tree.]{
        \includegraphics[width=0.3\textwidth,height=4.5cm]{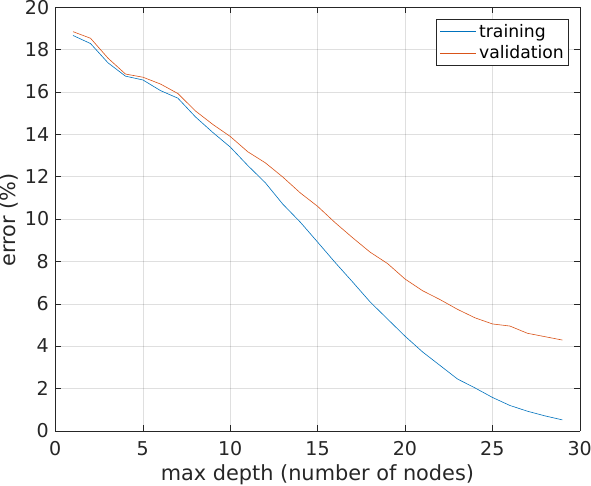}
        \label{fig:decisionTreeClas}
	}    

    \end{tabular}
    \caption{Accuracies on the training and the validation set for different max depth values when predicting lane merges.}
    \label{fig:Maxdepthmerges}
\end{figure*}

In this section, \textit{true} and \textit{false} recommendations for lane merges are analysed. The classifiers used for the predictions are selected from the most common machine learning algorithms. The learning phase helped us select a smaller number of algorithms that behaved the best out of the initial selection. This also allowed us to consider a smaller set of classifiers with greater granularity. For each of the classifiers, different parameters were considered. In order to select the most suitable algorithm for predicting lane merges, $9$ different algorithms were trained and validated. Whether a merge is predicted as \textit{true} or \textit{false}, a prediction for acceleration and heading is also provided. Predicting acceleration and heading involves a higher level of complexity. A few recommendations for acceleration and heading might be correct at the same time. For this reason, $3$ algorithms are presented in each case as they provide the best-obtained results. 

\subsection{Predicting lane merges}

To avoid over-fitting with default attributes, the parameters for maximum depth and number of estimators were assigned, since they specify the depth of the tree and the number of trees in the forest. These were selected by picking the values which were not over-fitting the model. We have selected the three highest scored algorithms to demonstrate how maximum depth was selected. Fig. \ref{fig:Maxdepthmerges} shows the accuracy for Random Forest, K-Nearest Neighbours and Decision Tree; which have the highest scores. To select the maximum depth of the tree, $30$ consecutive depths were considered for Random Forest and Decision Tree, and $50$ for K-Nearest Neighbours. The values considered for the number of estimators were: $1$, $2$, $5$, $10$, $20$, $35$, $50$, $75$ and $100$. The best value for the number of estimators is $100$; it is used for the three algorithms.

To select a proper maximum depth for the Random Forest, we estimated the error on the validation and training sets (Fig. \ref{fig:randomForestRecFeatures}) for different values of maximum depth. The value of maximum depth was chosen by comparing the results obtained on the validation set with the specified number for depth. The value of $16$ was the last for which the accuracy of the validation set was not worse than $1.5\%$ than the accuracy on the training set. The accuracy on the test set was the same as the accuracy on the validation set. The same technique was used to train the no over-fitting Decision Tree (Fig. \ref{fig:decisionTreeClas}). The best value for which the model was not over-fitting was equal to $11$ and showed less than $1\%$ of difference with the validation set. The over-fitting of K-Nearest Neighbours has a different property (Fig. \ref{fig:knnClass}). The lower the number of $K$, the higher the probability of over-fitting. To minimize the chance of over-fitting, $K$ was set to $50$. Table \ref{table:ResultsML} shows a summary of the obtained scores for predicting lane merges.

The scores were calculated using two functions. The function names are \textbf{\emph{accuracy\_score}} and \textbf{\emph{score}}. The \textbf{\emph{accuracy\_score}} function (sklearn.metrics) uses multi-label classification. This method calculates subset accuracy such that the predicted labels for a sample specifically equal a set of labels; the expected and correct labels. Crucially, this function was used to calculate the scores on the validation data set. The \textbf{\emph{score}} function returns the mean accuracy on the given test data and labels. In a multi-label classification environment, this is the subset accuracy that enforces a harsh metric. For each sample, it is required that every label set is correctly predicted. Crucially, this function was used to calculate scores on the training data set.

Additionally, for the Random Forest Classifier and Random Forest Regressor, we used the \textbf{\emph{cross\_val\_score}} function (sklearn.model\_selection). This function evaluates a score using cross-validation techniques. The parameter, \emph{cv}, was assigned a value of 10, which specifies the number of folds in a StratifiedKFold. Finally, results for both data sets were rounded to two decimal places.

\begin{table}[!t]
	\caption{Scores of different machine learning algorithms when predicting lane merges, Acceleration and Heading.}
    \label{table:ResultsML}
    \centering
    \scalebox{1}[1]{
    \begin{tabular}{lccc}
    	 \toprule
         \toprule
         Model & Merge & Acceleration & Heading \\
         \midrule

         Random Forest & $90.87$ & $75.74$ & $61.20$ \\
         K-Nearest Neighbours & $87.05$ & $-$ & $-$ \\
         Decision Tree & $86.84$ & $-$ & $-$ \\
         Gradient Boosting Classifier & $84.22$ & $-$ & $-$ \\
         Stochastic Gradient Decent & $80.54$ & $-$ & $-$ \\
         Logistic Regression & $80.41$ & $-$ & $-$ \\
         Linear SVC & $80.41$ & $-$ & $-$ \\
         Naive Bayes & $72.00$ & $-$ & $-$ \\
         Perceptron & $19.59$ & $-$ & $-$ \\
         Gradient Boosting & $-$ & $76.55$ & $62.85$ \\
         Linear Regression & $-$ & $15.47$ & $41.71$ \\
         \bottomrule
         \bottomrule
    \end{tabular} }
\end{table}

\subsection{Predicting acceleration}

By using the same data for the lane merge recommendations as the predictions of heading and acceleration, the results gave very low accuracy. For this reason, two simplifications on the dataset were made. Firstly, for each feature individually, a rounding of values was applied (e.g., lengths of the vehicles were rounded to $1$ decimal place). Secondly, the recommended acceleration was rounded to the closest integer. Furthermore, a function labeling predictions to be \textit{true} or \textit{false} by assumption that the result is only $1$ $\frac{m}{s^2}$ away from the correct value is correct, was implemented and used. 

\begin{figure}[!t]
    \centering
    \vspace{0.02in}
    \begin{tabular}{c}
    
    \subfloat[Gradient Boosting.]{
        \includegraphics[width=0.9\columnwidth]{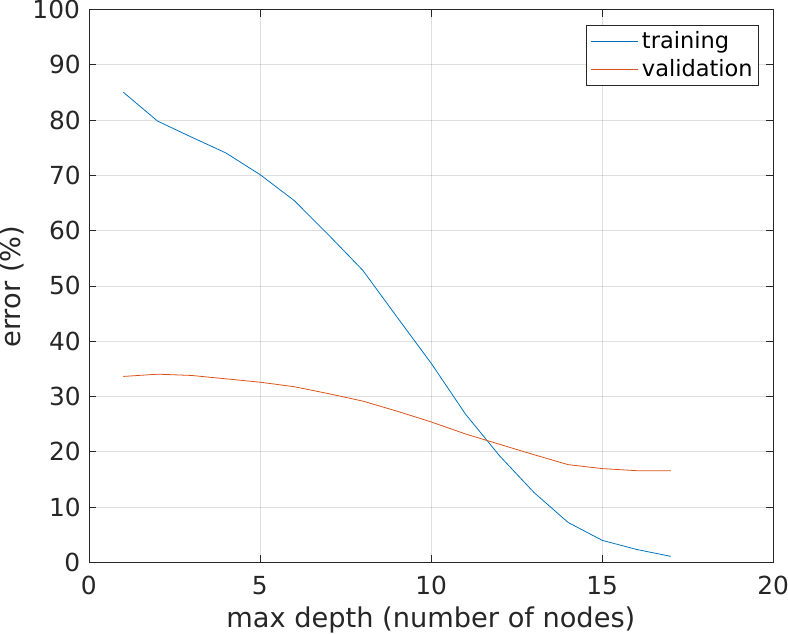}
        \label{fig:TetsValBoosAcc}
	} \\
    \subfloat[Random Forest.]{
        \includegraphics[width=0.9\columnwidth]{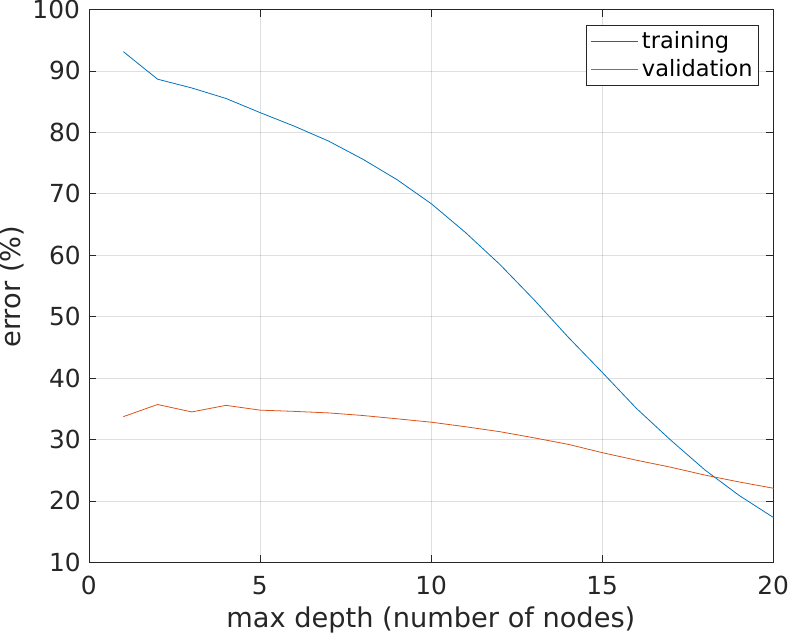}
        \label{fig:randReg}
	}

    \end{tabular}
    \caption{Accuracies on the training and the validation set for different max depth values when predicting acceleration.}
    \label{fig:Maxdepthacceleration}
\end{figure}

As previously mentioned, to avoid over-fitting, several tests were deployed with different values for the number of estimators and maximum depth. Again, the best output was when the number of estimators was set to $100$. We have selected the two highest scored algorithms to demonstrate how maximum depth was selected (Fig. \ref{fig:Maxdepthacceleration}). The considered values for maximum depth were in a range from $1$ to $13$ inclusively for Gradient Boosting (Fig. \ref{fig:TetsValBoosAcc}). The model with a maximum depth bigger than $11$ had much more accurate predictions for the training set than for the validation set, which suggested over-fitting. For this reason, a value of $11$ was selected for maximum depth. For Random Forest, a range between $1$ and $22$ was considered for maximum depth; the results are shown in Fig. \ref{fig:randReg}. The plot suggests that the values of maximum depth that are bigger than $18$ are getting over-fitted. Table \ref{table:ResultsML} shows a summary of the obtained scores for predicting acceleration. 

\subsection{Predicting heading}

Predicting heading is similar to predicting acceleration, where at the same time a few predictions might be true. However, a new consideration needs to be taken into account: two different values can lead to the same point, for instance, $10\degree$ and $-350\degree$. We applied a few simplifications as we did for acceleration: a) rounding the number to at most $1$ decimal place for non-location variables and b) rounding to $2$ decimal places for positioning information. Additionally, an evaluation function for heading was implemented, which returns values in a range between $0\degree$ and $360\degree$. It also calculates maximum error by comparing predicted and labelled values and calculating their difference.

\begin{figure}[!t]
    \centering
    \vspace{0.02in}
    \begin{tabular}{c}
    
    \subfloat[Gradient Boosting.]{
        \includegraphics[width=0.9\columnwidth]{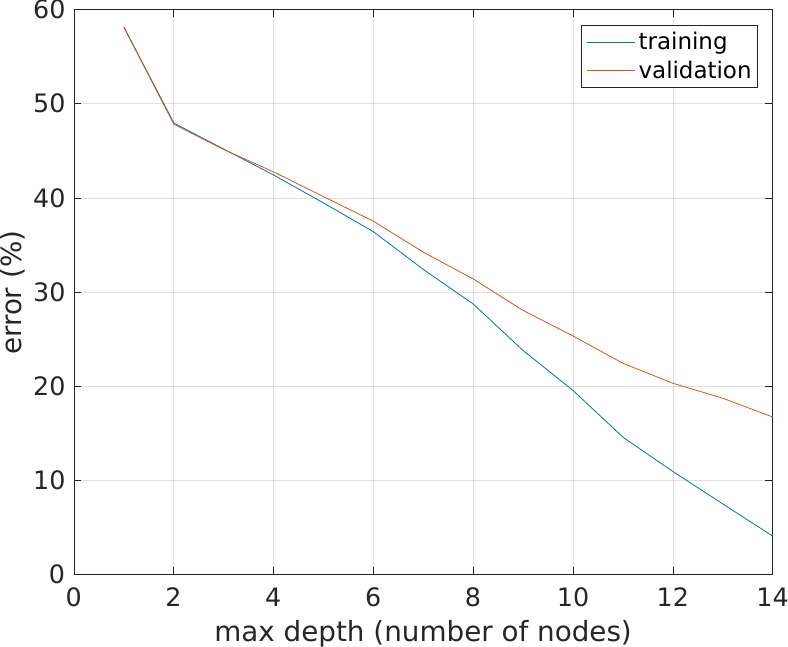}
        \label{fig:gradientBoHead}
	} \\
    \subfloat[Random Forest.]{
        \includegraphics[width=0.9\columnwidth]{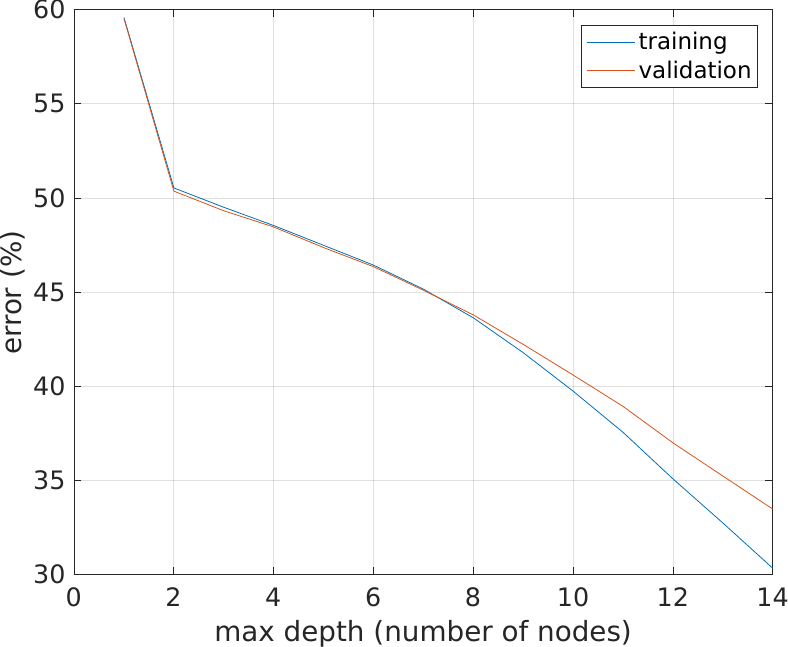}
        \label{fig:randomForestHead}
	}

    \end{tabular}
    \caption{Accuracies on the training and the validation set for the different max depth values when predicting heading.}
    \label{fig:Maxdepthheading}
\end{figure}

Following the same procedure as for acceleration, the best results were obtained when the number of estimators was set to $100$. We have selected the two highest scored algorithms for showing how maximum depth was selected (Fig. \ref{fig:Maxdepthheading}). To select maximum depth, tests were repeated from $1$ to $17$ for Gradient Boosting (see Fig. \ref{fig:gradientBoHead}) and from $1$ to $17$ for Random Forest (see Fig. \ref{fig:randomForestHead}). The value of the maximum depth for the Gradient Boosting was assigned to $6$ because, as for the previous cases, it is the last value for which the score of the validation set is not much worse than the score for the training set. The maximum depth for the Random Forest was set to $11$. The score of the algorithms for predicting heading is not very high (Table \ref{table:ResultsML}), but even after ignoring over-fitting, the model had a problem to learn anything efficiently. The score of the Random Forest is only $2\%$ worse on the validation set than the Gradient Boosting. 

\section{Conclusion} 
\label{Conclusion}
In this paper, we present a lane merge coordination model based on a centralised system. The system delivers trajectory recommendations to connected vehicles on the road. Real data from two highways has been used to train and validate a multitude of machine learning algorithms. Several tests were performed to properly select the maximum depth and the number of estimators for each algorithm. The Random Forest approach has been identified, among $9$ different algorithms, as the main candidate to predict \textit{true/false} lane merges. Predicting acceleration and heading involves a higher level of error and complexity. This is due to a scarcity of predicted values that fit the problem domain at the time of analysis. 

When predicting lane merges, the proposed coordination model presents meaningful results. On the other hand, predictions for acceleration and heading showed less accurate results. The error rate can have serious negative ramifications if handled inappropriately. A way to decrease the number of potential accidents caused by incorrectly labelled outputs is to run the predictive model, for each lane change, multiple times during the lane changing process. For this, a performance analysis of the model that is relying on communication standards is recommended as a future work. Therefore, an analysis of the network reliability, end-to-end latency measurements, combined with the computational analysis of the suggested algorithms and environment-based validation checks, will help this model gain more traction in a real-world scenario.

\section*{Acknowledgement}

This work has been performed in the framework of the H2020 project 5GCAR co-funded by the EU. The views expressed are those of the authors and do not necessarily represent the project. The consortium is not liable for any use that may be made of any of the information contained therein. This work is also partially funded by the EPSRC INITIATE and The UK Programmable Fixed and Mobile Internet Infrastructure.

\ifCLASSOPTIONcaptionsoff
  \newpage
\fi



\bibliographystyle{IEEEtran}
\bibliography{references}
\end{document}